\documentclass[balance,upint,subscriptcorrection,varvw,mathalfa=cal=boondoxo,pdf-a,colorlinks]{asmeconf}

\hypersetup{%
	pdfauthor={Giuseppe Del Giudice},
	pdftitle={Design Considerations for 3RRR Parallel Robots with Lightweight Approximate Static-Balancing},           
	pdfkeywords={Surgical robotics, parallel robots, statically balanced manipulators},
	pdfsubject = {IDETC-CIE 2023 Paper},
}

\ConfName{Proceedings of the ASME 2023\linebreak International Design Engineering Technical Conferences \& \linebreak Computers and Information in Engineering Conference}
\ConfAcronym{IDETC-CIE 2023}
\ConfDate{August 20–23, 2023} 
\ConfCity{Boston, MA} 
\PaperNo{IDETC2023-110947}

\usepackage{balance}  
\usepackage{multirow}
\usepackage{graphics}
\usepackage{floatrow}
\usepackage{times}
\usepackage{amsmath}
\usepackage{wrapfig}
\usepackage{multirow}
\usepackage{amsfonts}
\usepackage{leftidx}
\usepackage{indentfirst}
\usepackage{lipsum}
\usepackage{blkarray, bigstrut}
\usepackage{epsfig}
\usepackage{prettyref}
\usepackage{colortbl}
\usepackage[e]{esvect} 
\usepackage{graphicx}
\usepackage{caption}
\usepackage{subcaption}
\usepackage{algorithm}
\usepackage{algpseudocode}

\usepackage{arydshln} 
\usepackage{hyperref} 
\usepackage{natbib}
\usepackage{cases}
\usepackage{fancyhdr}

\makeatletter
\newcommand*{\algrule}[1][\algorithmicindent]{\makebox[#1][l]{\hspace*{.5em}\vrule height .75\baselineskip depth .25\baselineskip}}%

\newcount\ALG@printindent@tempcnta
\def\ALG@printindent{%
    \ifnum \theALG@nested>0
        \ifx\ALG@text\ALG@x@notext
            \addvspace{-3pt}
        \else
            \unskip
            \ALG@printindent@tempcnta=1
            \loop
                \algrule[\csname ALG@ind@\the\ALG@printindent@tempcnta\endcsname]%
                \advance \ALG@printindent@tempcnta 1
            \ifnum \ALG@printindent@tempcnta<\numexpr\theALG@nested+1\relax
            \repeat
        \fi
    \fi
    }%
\usepackage{etoolbox}
\patchcmd{\ALG@doentity}{\noindent\hskip\ALG@tlm}{\ALG@printindent}{}{\errmessage{failed to patch}}
\makeatother

\usepackage{booktabs}
\usepackage{tikz}

\newcommand{\realfield}{\hbox{I \kern -.4em R}}
\newcommand {\mb}[1]{\mathbf{#1}} 
\newcommand {\bs}[1]{\boldsymbol{#1}}
\newcommand{\uvec}[1]{\hat{\mathbf{#1}}}

\newcommand{\T}{^{\mathrm{T}}}  
\newcommand{\Rot}[2]{{^{#1}\mathbf{R}}_{#2}}  
\newcommand{\half}[1]{\frac{#1}{2}}
\newcommand*\circled[1]
{\tikz[baseline=(char.base)]{\node[circle,minimum size=7pt,draw=black,inner sep=0.5pt](char){\scriptsize #1};}}

\usepackage{leftidx}
\usepackage{mathtools}
\newif\ifTrackChanges   
\TrackChangesfalse 

\usepackage{cancel} 
\usepackage{xcolor}  
\definecolor{ao(english)}{rgb}{0.0, 0.5, 0.0}  

\usepackage[textsize=scriptsize, bordercolor=white,backgroundcolor=gray!30,linecolor=black,colorinlistoftodos]{todonotes}  
\usepackage[normalem]{ulem} 
\usepackage{soul} 

\ifTrackChanges
    \definecolor{tablecorr}{rgb}{1.0,0.0,0.0}   
    
    \newcommand{\remindlab}[2]{{[\colorbox{cyan}{#1}]}{\color{blue}{#2}}}
    \newcommand{\corr}[1]{{\color{red}{#1}}}
    \newcommand{\corrlab}[2]{{[\colorbox{yellow}{#1}]~}{\color{red}{#2}}}
    \newcommand{\cut}[1]{{\color{gray}{#1}}}
    \newcommand{\cutlab}[2]{{\colorbox{yellow}{[#1]}~}{{\color{gray}#2}}}
    \newcommand{\cutEq}[2]{\begin{equation}\label{#2}\color{gray}{\xcancel{#1}}\end{equation}}   
    \newcommand{\rewrite}[1]{{\color{ao(english)}{#1}}}
    \newcommand{\rewritelab}[2]{{\colorbox{ao(english)}{[#1]}~}{\color{ao(english)}{#2}}}
\else
    
    \definecolor{tablecorr}{rgb}{0.0,0.0,0.0}  
    \newcommand{\remindlab}[2]{{{#2}}}
    \newcommand{\corr}[1]{{{#1}}}
    \newcommand{\corrlab}[2]{{{#2}}}
    \newcommand{\cut}[1]{{}}
    \newcommand{\cutlab}[2]{{}}
    \newcommand{\cutEq}[2]{}
    \newcommand{\rewrite}[1]{{#1}}
    \newcommand{\rewritelab}[2]{{#2}}
\fi

\usepackage{multirow}  
\usepackage{tabularx}  
\usepackage{colortbl} 
\usepackage{footnote}
\makeatletter
\newcommand{\thickhline}{%
    \noalign {\ifnum 0=`}\fi \hrule height 2pt
    \futurelet \reserved@a \@xhline
}
\newcolumntype{"}{@{\hskip\tabcolsep\vrule width 2pt\hskip\tabcolsep}}
\makeatother
\usepackage{enumerate}
\usepackage{enumitem}
\usepackage{pifont}   
\newcommand{\arrowbullet}{\par \noindent \ding{227}~}

\begin{document}
\pagestyle{fancy}
\fancyhf{} 
\renewcommand{\headrulewidth}{0pt}
\lhead{ASME IDTEC-CIE 2023. Accepted Version.}
\title{Design Considerations for 3RRR Parallel Robots\\ with Lightweight, Approximate Static-Balancing}
%

%
%
%

\SetAuthors{%
	Giuseppe Del Giudice\affil{1}, 
	Garrison L.H. Johnston\affil{1}, 
	Nabil Simaan\affil{1}\CorrespondingAuthor{nabil.simaan@vanderbilt.edu}
	}

\SetAffiliation{1}{Dept. of Mechanical Engineering, Vanderbilt University, Nashville TN, USA}

\maketitle
\begin{abstract}
Balancing parallel robots throughout their workspace while avoiding the use of balancing masses and respecting design practicality constraints is difficult. Medical robots demand such compact and lightweight designs. This paper considers the difficult task of achieving optimal approximate balancing of a parallel robot throughout a desired task-based dexterous workspace using balancing springs only. While it is possible to achieve perfect balancing in a path, only approximate balancing may be achieved without the addition of balancing masses. Design considerations for optimal robot base placement and the effects of placement of torsional balancing springs are presented. Using a modal representation for the balancing torque requirements, we use recent results on the design of wire-wrapped cam mechanisms to achieve balancing throughout a task-based workspace. A simulation study shows that robot base placement can have a detrimental effect on the attainability of a practical design solution for static balancing. We also show that optimal balancing using torsional springs is best achieved when all springs are at the actuated joints and that the wire-wrapped cam design can significantly improve the performance of static balancing. The methodology presented in this paper provides practical design solutions that yield simple, lightweight and compact designs suitable for medical applications where such traits are paramount.
\end{abstract}


\section{Introduction}\label{ch:intro}

\par Parallel mechanisms have gained use in several applications (e.g.\corrlab{R4.21}{,}  rapid prototyping machines, fast pick-and-place robots, haptic devices and surgical robots) due to their potential benefits of precision, compactness, stiffness and dynamic agility.\cutlab{R2.2}{When attempting to achieve static balancing for these mechanisms, the design task is complicated by the presence of passive and active joints in their kinematic chains.} Unlike serial robots where exact static balancing of each joint may be directly decoupled as a sub-problem to be solved by judicious placement of energy storage elements (e.g.\corrlab{R4.21}{,}  \cite{morita2003parallelogram_balancing,Lin2010_serial_spring_balance,Kim2013}), such problem decomposition is \cut{hard}\rewrite{harder} to achieve in parallel mechanisms because of the coupled nature of all the closed loop kinematic chains of these mechanisms.
\par The motivation for statically balanced parallel robots stems from reducing actuator torque requirements and enabling compactness and miniaturization. Because statically balanced robots require smaller actuation torques, they offer safer human-robot interaction \cite{Vermeulen2010,Whitney2014}. Perfectly balanced mechanisms allow free motion without motors/brakes to keep their position. For example, statically balanced mechanisms have been used for offsetting the physical effort of manufacturing workers in tasks involving payload transfer \cite{Lacasse2013_hydraulic}, for offering assistance to patients with weak limbs \cite{fattah2006sit_stand,Lin2013_arm_balance} and for supporting surgical or medical imaging instruments \cite{Lessard2007a,Seo2014_surgical_balanced}.
\par  There have been three means for achieving static balancing of parallel mechanisms: a) using counterbalancing masses (e.g.\corrlab{R4.21}{,}  \cite{Foucault2004,Lecourse2010}) , b) using springs for energy storage (e.g.\corrlab{R4.21}{,}  \cite{Kilic2012,Simionescu2011}), and c) using both springs and counterbalance masses (e.g.\corrlab{R4.21}{,} \cite{Laliberte1999_springs_and_masses,Kang_yi_parallelogram_balancing2016}). \remindlab{R1.2}{We limit this work to solutions using only springs for static balancing due to the drawbacks of counterbalance masses as manifested in increased weight and bulk of the overall parallel mechanism}. Spring-based design solutions are limited to applications where the direction of the gravity vector does not change relative to the base of the parallel linkage.
\par For actuated parallel robots, the requirement for exact static balancing is not as strict as for passive statically-balanced mechanisms. Therefore, in this paper, we propose a design strategy for approximate static balancing of parallel robots that retains the design simplicity of the kinematic chains by only using torsional springs or wire-wrapped cams.
\par Although there have been many works in the area of static balancing of parallel linkages, many of the designs proposed are too cumbersome to be attractive for a realistic embodiment. Exceptions to this observation are \cite{ebert2000pantograph,ebert2002pantograph_balance} who proposed and demonstrated a practical embodiment of a six degrees of freedom (DOF) parallel mechanism with spring-balanced parallelogram linkages. \corrlab{R2.1}{It was shown in \cite{ebert2000pantograph} that the use of three spring-balanced, parallelogram-based linkages as the supporting legs of a parallel robot results in relatively strict conditions on the location of the center of mass of the payload in order to achieve static balancing. These constraints may be relaxed by adding at least four kinematic chains to support a moving platform; therefore, resulting in cumbersome designs. Examples of six DOF parallel robots supported by six kinematic chains were also presented in \cite{Gosselin2000DOF_balancing_springs}, but also relied on parallelograms as part of each kinematic chain - thereby increasing the mechanical complexity and hindering miniaturization}.
\par This paper presents a practical design alternative for achieving statically balanced parallel mechanisms. Although exact balancing throughout a workspace may be achieved in particular geometries with the addition of balancing masses \corr{or the use of multiple kinematic chains with parallelogram linkages}, we avoid the use of such balancing masses \corr{or linkages} in favor of \corr{retaining} a compact and lightweight design. This design choice is motivated by our envisioned target applications of portable and compact parallel robots for assistive applications in surgery (e.g.\corrlab{R4.21}{,} cooperative robots for tremor filtering \cite{TaylorSteady_hand1999}) or for manipulation of a fixed payload (e.g.\corrlab{R4.21}{,} remote ultrasound sonography within an MRI bore\cite{Lessard2007a}). The advantage of compactness and design practicality sought in this paper comes at the price of achieving only approximate balancing throughout a desired dexterous workspace as opposed to exact balancing. 
\par Because of the target application domain, we avoid the use of non-encapsulated spring elements (e.g., a linear spring or a wire passing in the air and connecting two links in the mechanism). Such designs offer a risk of injury by potentially grabbing tissue and complicating the task of sterilization using sterile drapes that may also be caught on these springs. With these design constraints in mind, we focus this paper on the 3RRR parallel robot architecture because of its simplicity and relevance to medical applications (e.g. \cite{Pile2014_3RRR_cochlea}). 
\par
\begin{figure}[htbp]
\includegraphics[width=0.7\columnwidth]{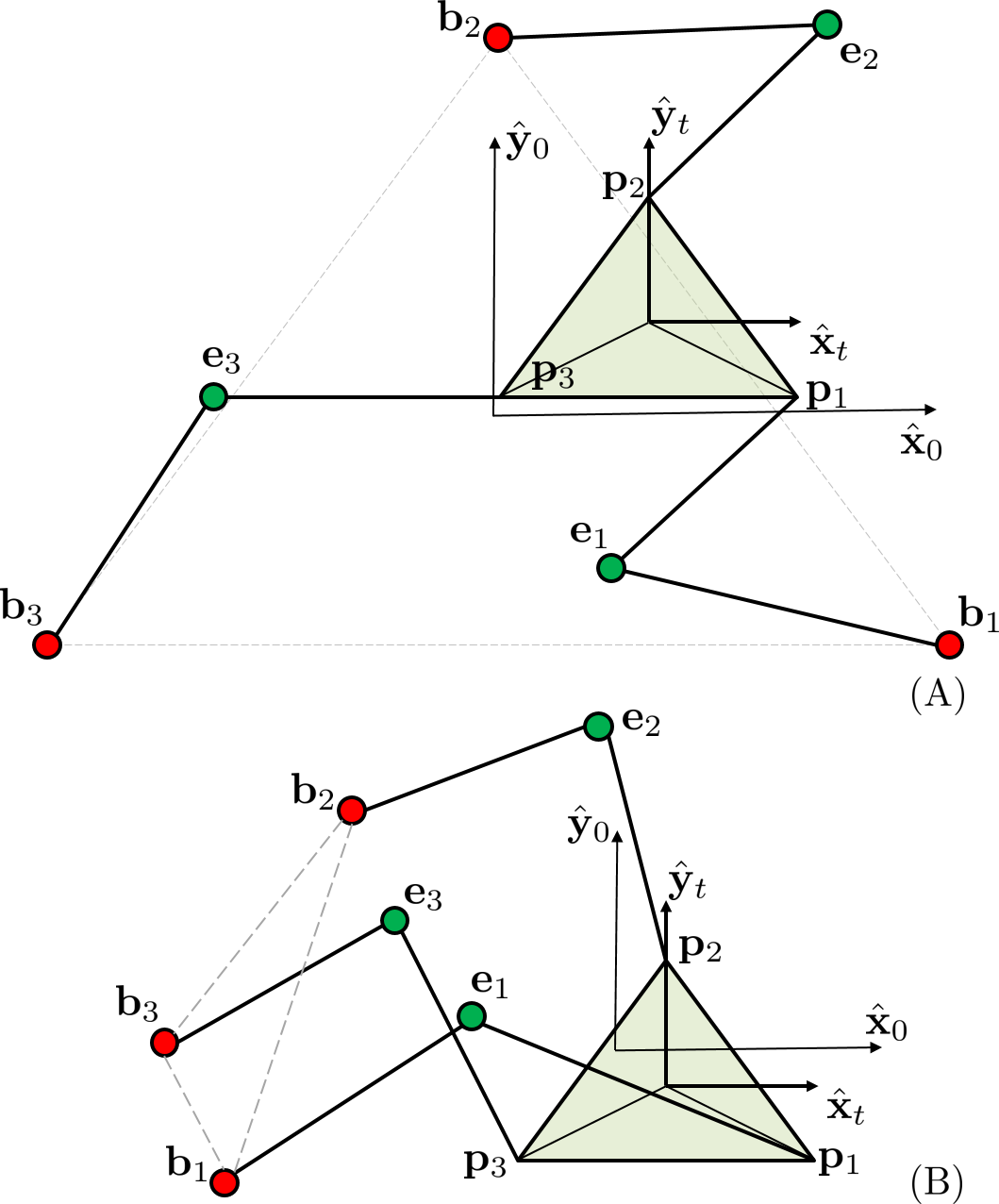}
\caption{3-RRR Planar manipulator: A) wide (WL) layout B) Narrow (NL) layout. The red and green dots designate the active and passive joints, respectively.}
\label{fig:3RRR_combined}
\end{figure}

 With these sought design traits, we consider two design alternatives: 1) parallel mechanisms using torsional springs in either their active and/or in their passive joints, 2) parallel mechanisms using a combination of linear springs at the base with wire-wrapped cams at the base. This design option is inspired by Koser \cite{koser2009cam} who presented a cam-balancing mechanism for serial robots and by Kilic et al. \cite{Kilic2012} who presented a wire-wrapped cam mechanism for achieving adjustable stiffness.
In addition to the above design alternatives for balancing, We also consider the effect of the chosen layout of the 3RRR robot architecture, which may be configured with a wide or a narrow layout as shown in Fig.~\ref{fig:3RRR_combined} where the moving platform is contained within/outside the convex hull of the base joints, respectively.

\par Given that we can achieve only approximate balancing for a desired workspace, this paper presents a simulation study aimed at answering the following questions:  
\begin{itemize}[label=\ding{227},noitemsep,labelindent=0cm]
    \item What are the best locations for the placement of balancing springs along a 3RRR kinematic chain?
    \item To what extent does mechanical layout (wide v.s. narrow) affect the feasibility of static balancing using torsional springs?. 
    \item Another option to achieve static balancing is to use wire-wrapped cams as in \cite{koser2009cam,Kilic2012,Johnston2023_wire_wrapped_cams} at the base joints. To what extent is the simplicity of balancing using torsional springs at the base is justified compared to the alternative of using wire-wrapped cams? 
\end{itemize}
\par Relative to prior literature, we present a contribution that uses the 3RRR parallel robot architecture as a sample application for our design strategy and we illustrate the importance of selecting the possible locations for the balancing springs by addressing the questions listed above.  We also present an approach whereby approximate balancing is achieved throughout a portion of the workspace by deriving a modal approximation of the desired balancing spring torque in a wire-wrapped cam mechanism and adapting the results to produce possible cam implementations using Kilic's closed-form approach in \cite{Kilic2012}. We believe that this exploration will be useful for guiding the design process for practical compact and lightweight parallel mechanisms for the applications mentioned above.  
\section{Static Balancing Using Spring Elements}
We consider the scenario where $\uvec{g}$ designates a unit vector in the direction of gravity and $g$ designates the acceleration of gravity. We assume a parallel robot subject to an external force $\mb{f}_e$ and a moment $\mb{m}_e$. With this robot having the set of active joints $\mb{q}$ and the set of passive joints $\bs{\phi}$, we use $\bs{\tau}$ to designate the actuator torques. We also use $\mb{x}$ to denote the pose of the moving platform which is defined as the vector of position and orientation parametrization. For example, $\mb{x}$ for a planar robot would be the two Cartesian coordinates and the angle of rotation of the moving platform.
\par Loop closure constraint equations of the topologically independent loops of a parallel robot may be used to obtain the necessary nonlinear equations relating the passive joints $\bs{\phi}$ to the active joints $\mb{q}$. By taking a time derivative of these loop closure constraint equations, one can always obtain this instantaneous internal kinematics mapping:
\begin{equation}\label{ab_vector}
  \mb{A}\dot{\bs{\phi}} = \mb{B}\dot{\mb{q}}
\end{equation}
And the Jacobian\footnote{we use $\mb{J}_{xy}$ to designate the Jacobian such that $\dot{\mb{x}}=\mb{J}_{xy}\dot{\mb{y}}$.} associated with the above mapping can be defined as:
\begin{equation}\label{j_phi_q_def}
  \dot{\bs{\phi}}=\mb{J}_{\phi q}\dot{\mb{q}}\quad ,\quad \mb{J}_{\phi q} = \mb{A}^{-1}\mb{B}
\end{equation}
Examples of this approach can be found in \cite{paul1986kinematics,simaan1999analysis} for planar and spatial parallel mechanisms.
\par Next, we consider the potential energy $V_g$ of the parallel mechanism. With $m_i$ designating the mass of the $i^{th}$ link, $h_{ci}$ denoting the height of its center of mass, $m_p$ the mass of the moving platform and $\mb{t}$ the location of the moving platform center (end effector point),  one can write:
\begin{equation}\label{eq:general_pot_energy}
  V_g = \sum_{i=1}^{n}\left(m_igh_{ci}\left(\mb{q},\bs{\phi}\right) + m_p\,g\left(\mb{t}^T\uvec{g}\right)\right)
\end{equation}
where $n$ designates the total number of links in the legs, excluding the base and moving platform.
\par Considering the case where a robot is subject only to gravitational forces, one can write the conditions for static equilibrium as:
\begin{equation}\label{eq:tau_g_without_load}
  \bs{\tau}_g^T d\mb{q} - d V_g = 0 \Rightarrow \bs{\tau}_g = \frac{dV_g}{d\mb{q}}
\end{equation}
In the above equation,\corrlab{R2.5}{where $\bs{\tau}_g$ is defined as the gravitational torque}, obtaining $V_g$ only as a function of the active joints $\mb{q}$ requires solving a set of nonlinear equations tantamount to solving the direct kinematics or the nonlinear loop closure equations.  Instead, we consider $V_g=V_g(\mb{q},\bs{\phi},\mb{x})$ and we apply the chain rule:

\begin{equation}\label{eq:tau_g_noload}
  \bs{\tau}_g=\frac{\partial V_g}{\partial \mb{q}} + \mb{J}_{\phi q}\T\frac{\partial V_g}{\partial\bs{\phi}} + \mb{J}_{xq}\T\frac{\partial V_g}{\partial \mb{x}}
  \end{equation}
\par In the general case of a robot having flexible energy storage elements and subject to an external wrench $\mb{w}_e$, the virtual work principle leads to:
\begin{equation}\label{eq:tau_g_with_extermal_load_and_springs}
  \bs{\tau}^T \delta\mb{q} - \delta V_g - \delta V_e - \mb{w}_e\T\delta \mb{x} = 0
\end{equation}
where $\delta V_e$ is the change of elastic potential energy that is associated with a virtual displacement $\delta \mb{q}$ and $\mb{w}_e\T\delta \mb{x}$ denotes the work of \cutlab{R1.5}{the robot on the environment}\rewritelab{R1.5}{the environment on the robot}.  \cutlab{R2.5}{By defining the gravitational torque $\bs{\tau}_g=\frac{dV_g}{d\mb{q}}$}\rewritelab{R2.5}{Using the definition for $\bs{\tau}_g$} and the torque due to the internal springs $\bs{\tau}_e=\frac{dV_e}{d\mb{q}}$, we obtain:
\begin{equation}\label{eq:tau_final_general_case}
  \bs{\tau} = \bs{\tau}_g + \bs{\tau}_e + \left[\mb{J}_{qx}^T\right]^{-1}\mb{w}_e
\end{equation}
Finally, by expressing the elastic energy $V_e$ as a function of all possible spring locations at the active and passive joints, we can write:
\begin{equation}\label{eq:potential_energy_def}
V_e=\half{1}\left(\tilde{\mb{q}}\T\mb{K}_q\tilde{\mb{q}} +
\tilde{\bs{\phi}}\T\mb{K}_\phi\tilde{\bs{\phi}}
\right)
\end{equation}
where $\tilde{\mb{q}}=\mb{q}-\mb{q}_f$ and $\tilde{\bs{\phi}}=\bs{\phi}-\bs{\phi}_f$ designate the spring deflections relative to their unloaded configurations with corresponding joint angles $\mb{q}_f$ and $\bs{\phi}_f$ and $\mb{K}_q$ and $\mb{K}_\phi$ are diagonal matrices containing the spring constants installed to the active and passive joints.
\par The torque due to the elastic elements $\bs{\tau}_e$ is given by:
\begin{equation}\label{eq:tau_e_def}
\bs{\tau}_e=\frac{d V_e}{d\mb{q}}=\frac{\partial V_e}{\partial \mb{q}} + \mb{J}_{\phi q}\T \frac{\partial V_e}{\partial \bs{\phi}}= \mb{K}_q\tilde{\bs{q}} + \mb{J}_{\phi q}^T\mb{K}_\phi\tilde{\bs{\phi}}
\end{equation}
\par Substituting  \eqref{eq:tau_e_def} into \eqref{eq:tau_final_general_case} gives the condition for static balancing for a given load $\mb{w}_e$:
\begin{equation}\label{eq:tau_final_general}
  \bs{\tau} =\bs{\tau}_g+ \mb{J}_{\phi q}^T\mb{K}_\phi\tilde{\bs{\phi}}+ \mb{K}_q\tilde{\bs{q}}+ \left[\mb{J}_{qx}^T\right]^{-1}\mb{w}_e=\mb{0}
\end{equation}
where $\bs{\tau}_g$ is given by \eqref{eq:tau_g_noload}.

\section{Planar 3-RRR Manipulator}\label{sec:3RRR}
\par To illustrate the approach of the previous section, we use a 3RRR planar parallel robot. We consider two possible layouts for this robot: the \textit{Wide  layout (WL)} shown in Fig.~\ref{fig:3RRR_combined}A and \textit{Narrow layout (NL)} shown in Fig.~\ref{fig:3RRR_combined}B. In the wide layout, the three base \cutlab{R1.6}{joint}\corrlab{R1.6}{joints} are placed radially with \corrlab{R1.6}{respect} to the world frame and equidistantly from each other. In the narrow layout, the three base joints are placed to the left of the world frame $\{\uvec{x}_0, \uvec{y}_0\}$ to mimic an hand-held device with its end effector reaching to the right. We also keep the serial kinematic chains in each leg in an elbow-up configuration to avoid the risk of collision between the elbows and the anatomy, which typically lies below the robot base.
\par In the following simulation study, we will investigate four possibilities for the placement of torsional springs:
\vspace{-0.5\baselineskip}
\begin{itemize}[noitemsep,leftmargin=0.3cm,labelindent=0cm]
  \item \textbf{Mode 0:} the baseline case where no springs are used. The actuator torque required for this case will be designated by $\bs{\tau}_0$.
  \item \textbf{Mode 1:} balancing springs are installed only at the active joints.
  \item \textbf{Mode 2:} balancing springs are installed only at the elbow joints.
  \item \textbf{Mode 3:} balancing springs are installed both at the active and elbow joints.
\end{itemize}
\corrlab{R1.4}{We note that we omit from this paper the case of using springs at the platform joints since our analysis has shown that the addition of these joints does not lead to actuator torque reduction and may even lead to increased actuator torque when the moving platform changes orientation.}
 \subsection{Kinematics}
\begin{figure}[htbp]
\includegraphics[width=0.6\columnwidth]{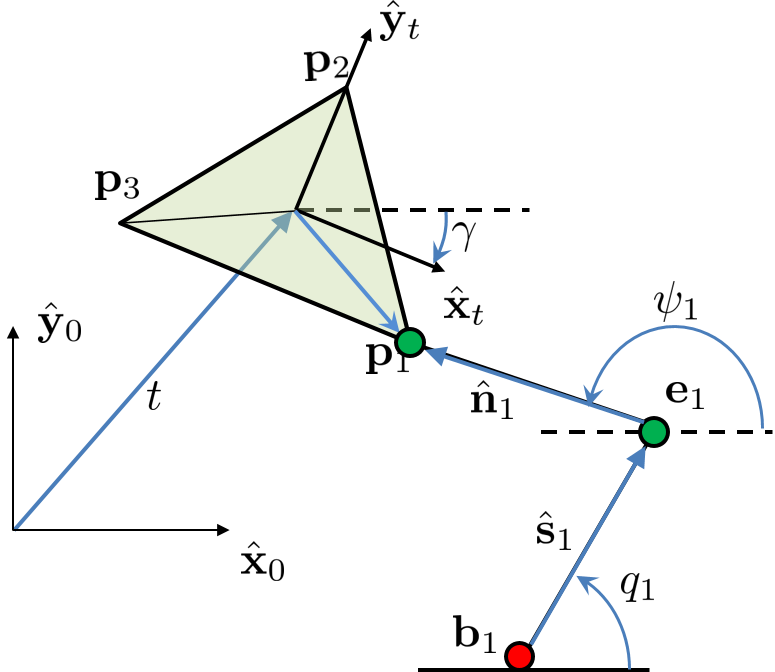}
\caption{Single kinematic chain of 3-RRR Parallel mechanism \label{fig:3RRR_chain_mechanism}}
\end{figure}

  Equation \eqref{eq:tau_final_general} requires the Jacobian matrices $\mb{J}_{xq}$ and $\mb{J}_{\phi q}$. We therefore derive these Jacobians in this section. Referring to Fig.~\ref{fig:3RRR_chain_mechanism}, we define $\mb{b}_i$ as the locations of the base joints in the base (world) frame $\{\uvec{x}_0, \uvec{y}_0\}$ and  ${}^m\mb{p}_i$, $i=1,2,3$  as the location of the corresponding passive joints in the moving platform-attached frame $\{\uvec{x}_t, \uvec{y}_t\}$\footnote{we use ${}^t\mb{x}$ to designate a vector described in frame ${\uvec{x}_t, \uvec{y}_t}$ and we omit the left superscript when the vector is described in world frame}. We also define $\uvec{s}_i$, as the unit vectors along the actively rotating links at the base and $\uvec{n}_i$ as the unit vectors along the forearms connecting the moving platform passive revolute joints at ${}^t\mb{p}_i$ to the elbow passive joints $\mb{e}_i$. Finally, we assume the leg lengths $||\mb{b}_i-\mb{e}_i||=L_1$ and $||\mb{p}_i-\mb{e}_i||=L_2$.
  %

%
\par Designating the rotation of the moving frame relative to the world frame as $\Rot{w}{t}$ and referring to Fig.~\ref{fig:3RRR_chain_mechanism}, we can write the loop closure equation \corrlab{R4.8}{for a single kinematic chain}:
\begin{equation}\label{eq:loop_closure}
\mb{t}+ \Rot{w}{t} {}^{t}\mb{p}_{1}-L_2\hat{\mb{n}}_1-L_1\hat{\mb{s}}_1-\mb{b}_1 = 0
\end{equation}
Taking the time derivative of \eqref{eq:loop_closure} and recalling that the time derivatives ${}^t\dot{\mb{p}}_1={}^w\bs{\omega}_t\times {}^t\mb{p}_1$, $\dot{\uvec{s}}_1=\dot{q}_1 \uvec{z}_0\times \uvec{s}_1$,  and \mbox{$\dot{\uvec{n}}_1=\dot{\phi}_1 \uvec{z}_0\times \uvec{n}_1$} we obtain:
\begin{equation}\label{eq:loop_deriv}
\dot{\mb{t}}+{}^w\bs{\omega}_{t}\times  \Rot{w}{t} {}^{t}\mb{p}_{1}-L_2\dot{\phi}_1\left(\hat{\mb{z}}_0\times \hat{\mb{n}}_1\right) - L_1\dot{q}_1\left(\hat{\mb{z}}_0\times \hat{\mb{s}}_1\right) = 0
\end{equation}
\par To find the relation between $\dot{\mb{q}}$ and task space twist, we define the end effector's reduced-dimension twist as \mbox{$\dot{\mb{x}}\triangleq [\dot{t}_x,\dot{t}_y,\dot{\gamma}]\T$} where the first two terms designate the Cartesian end effector speeds and the third element designates the angular rate of rotation of the moving platform. The passive joint speed $\dot{\phi}_1$ is next eliminated from the above equation by taking a dot product with $\uvec{n}_1$ and using the scalar triple product identity $\mb{a}\cdot \mb{b}\times \mb{c}=\mb{c}\cdot \mb{b}\times \mb{a}=\mb{b}\cdot \mb{c}\times \mb{a}$. These mathematical manipulations allow rearranging \eqref{eq:loop_deriv}  as:
\begin{equation}\label{eq:loop_definitiveq}
  \hat{\mb{n}}^T_1\dot{\mb{t}}+\left(\Rot{w}{t} {}^{t}\mb{p}_{1}\times \hat{\mb{n}}_1\right)^T \dot{\gamma}\uvec{z}_0 = \hat{\mb{n}}^T_1 \left(\hat{\mb{z}}_0 \times \hat{\mb{s}}_1\right)L_1\dot{q}_1
  \end{equation}
Since the 3RRR is a planar robot, $^{w}\bs{\omega}_t=\dot{\gamma}\uvec{z}_0$. Rewriting equation \eqref{eq:loop_definitiveq} with all three chains results in the matrix form:
\begin{equation}
\begin{split}
&
\underbrace{
\begin{bmatrix}
n_{1x} & n_{1y} & \uvec{z}_0\T(\Rot{w}{t}^{t}\mb{p}_1\times\uvec{n}_1) \\
n_{2x} & n_{2y} & \uvec{z}_0\T(\Rot{w}{t}^{t}\mb{p}_2\times\uvec{n}_2) \\
n_{3x} & n_{3y} & \uvec{z}_0\T(\Rot{w}{t}^{t}\mb{p}_3\times\uvec{n}_3)
\end{bmatrix}
}_{\mb{A}}
\underbrace{
\begin{bmatrix}
\dot{t}_x \\
\dot{t}_y \\
\dot{\gamma}
\end{bmatrix}}_{\dot{\bs{x}}}=\\
&
\underbrace{
L_1
    \begin{bmatrix}
        \hat{\mb{n}}^T_1 \left(\hat{\mb{z}}_0 \times \hat{\mb{s}}_1\right)& 0 & 0 \\
        0 & \hat{\mb{n}}^T_2 \left(\hat{\mb{z}}_0 \times \hat{\mb{s}}_2\right) & 0 \\
        0 & 0 & \hat{\mb{n}}^T_3 \left(\hat{\mb{z}}_0 \times \hat{\mb{s}}_3\right)
    \end{bmatrix}
}_{\mb{B}}
      \dot{\mb{q}}
\end{split}
\end{equation}
Using the above results, we define the instantaneous inverse kinematics Jacobian $\mb{J}_{qx}\triangleq \mb{B}^{-1}\mb{A}$ such that $\dot{\mb{q}}=\mb{J}_{qx}\dot{\bs{x}}$.

\par It is possible to extract the Jacobian matrix $\mb{J}_{\phi x}$ that relates $\dot{\mb{x}}$ to the passive joint rates $\dot{\bs{\phi}}$. To eliminate $\dot{q}_1$ from \eqref{eq:loop_deriv} we  take the dot product with $\hat{\mb{s}}_1$ and using again the cross product identity, which yields:
\begin{equation}\label{eq:loop_definitive_psi}
  \hat{\mb{s}}^T_1\dot{\mb{t}}+\left(\Rot{w}{t} {}^{t}\mb{P}_{1}\times \hat{\mb{s}}_1\right)^T \dot{\gamma}\uvec{z}_0 = \hat{\mb{s}}^T_1 \left(\hat{\mb{z}}_0 \times \hat{\mb{n}}_1\right)L_2\dot{\phi}_1
  \end{equation}
Repeating the \eqref{eq:loop_definitive_psi} for the three kinematic chains provides $\mb{J}_{\phi x}$ such that $\dot{\bs{\phi}}=\mb{J}_{\phi x}\dot{\mb{x}}$:
  \begin{equation}
\begin{split}
&
\underbrace{
\begin{bmatrix}
s_{1x} & s_{1y} & \uvec{z}_0\T(\Rot{w}{t}^{t}\mb{p}_1\times\uvec{s}_1) \\
s_{2x} & s_{2y} & \uvec{z}_0\T(\Rot{w}{t}^{t}\mb{p}_2\times\uvec{s}_2) \\
s_{3x} & s_{3y} & \uvec{z}_0\T(\Rot{w}{t}^{t}\mb{p}_3\times\uvec{s}_3)
\end{bmatrix}
}_{\mb{C}}
\underbrace{
\begin{bmatrix}
\dot{t}_x \\
\dot{t}_y \\
\dot{\gamma}
\end{bmatrix}}_{\dot{\bs{x}}}=\\
&
\underbrace{
L_2
    \begin{bmatrix}
        \hat{\mb{s}}^T_1 \left(\hat{\mb{z}}_0 \times \hat{\mb{n}}_1\right)& 0 & 0 \\
        0 & \hat{\mb{s}}^T_2 \left(\hat{\mb{z}}_0 \times \hat{\mb{n}}_2\right) & 0 \\
        0 & 0 & \hat{\mb{s}}^T_3 \left(\hat{\mb{z}}_0 \times \hat{\mb{n}}_3\right)
    \end{bmatrix}
}_{\mb{D}}
      \dot{\bs{\phi}}
\end{split}
\end{equation}
Where is given by $\mb{J}_{\phi x}=\mb{D}^{-1}\mb{C}$.
   \par Using the above two Jacobians, it is possible to derive $\mb{J}_{\phi q}$ such that
   $\dot{\bs{\phi}}=\mb{J}_{\phi q}\dot{\mb{q}}$ as the following:
 \begin{equation}\label{eq:J_phi_q}
     \dot{\bs{\phi}}=\mb{J}_{\phi x}\dot{\mb{x}}=\mb{J}_{\phi x}{\mb{J}_{qx}}^{-1}\dot{\mb{q}}\quad  \Rightarrow \quad \mb{J}_{\phi q}=\mb{J}_{\phi x}{\mb{J}_{qx}}^{-1}
\end{equation}
In the above derivations, we assume that the robot does not suffer any type of singularity associated degeneracy of any of the matrices $\mb{A}$, $\mb{B}$, $\mb{C}$, $\mb{D}$. Conditions for degeneracy of these matrices were discussed in the literature including \cite{Gosselin1990singularity3rrr,Xiang2012}.
\section{Static Balancing Strategy and Simulation Results}
\par Given a dexterous workspace, one can be tempted to design a static balancing solution for the entire dexterous workspace. We do not advocate for this because the solution obtained through this approach will have attenuated effectiveness. Attempting to obtain maximal static balancing throughout the dexterous workspace will demand non-monotonous balancing torque profiles with very large ranges that would effectively require using either an adjustable linkage to change the spring pre-load or incorporating balancing masses in addition to springs. In this section, we propose a different design strategy for simplified static balancing implementation by using a task-based design task and identifying the ideal location of the task of the robot.
\subsection{Task-Based Static Balancing and Sub-workspace Location}
\par
\begin{figure*}[h]
\includegraphics[width=0.99\columnwidth]{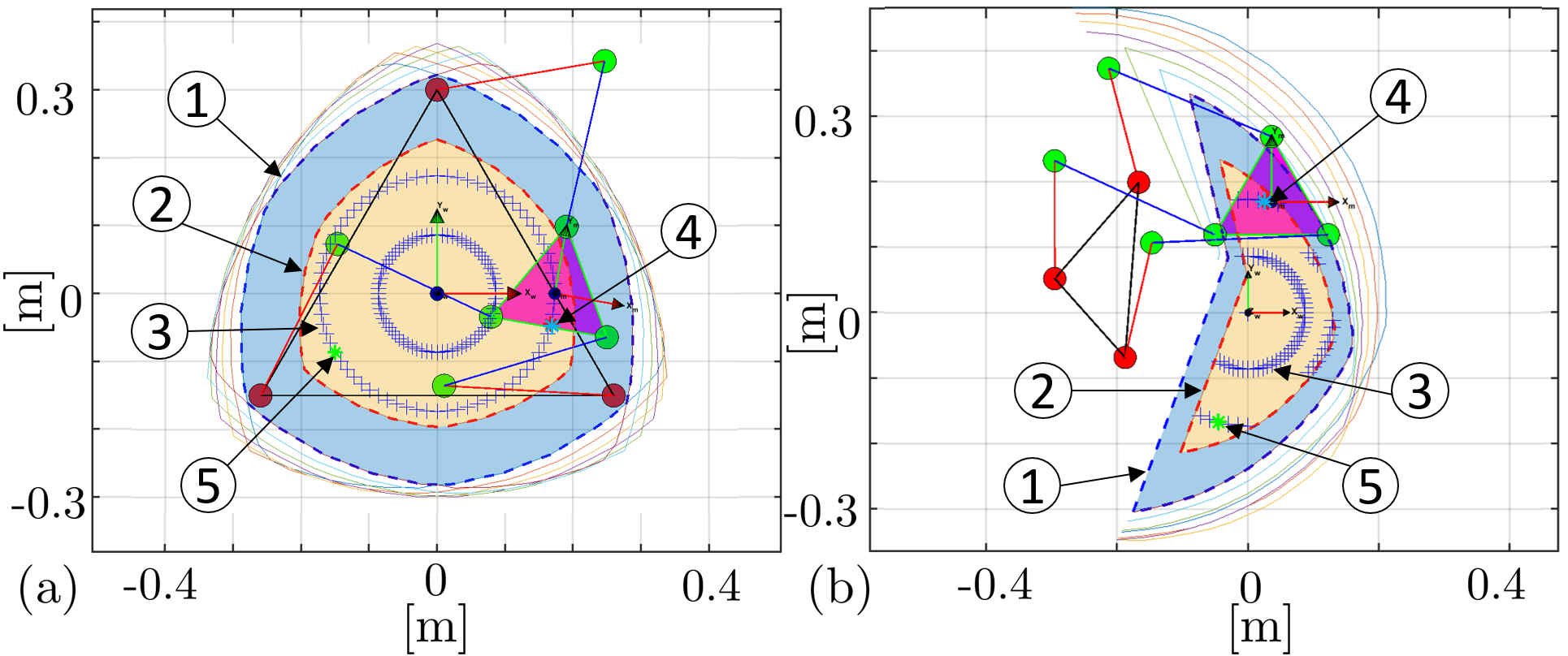}
\caption{A) WL robot configuration. B) NL robot configuration. Both manipulators are placed in the configuration of max percentage of torque reduction - \protect\circled{1} Dexterous Workspace \protect\circled{2} Dexterous sub-workspace \protect\circled{3} Scanned points for optimal location for the task-based dexterous sub-workspace \protect\circled{4}-\protect\circled{5} Point of maximal and minimal required torque respectively. \label{fig:optimization_configuration_pic}}
\end{figure*}

 The first step towards optimizing the static balancing solution for a given robot is the identification of the dexterous workspace within which the robot can satisfy desired ranges for Cartesian and rotation motions. Within the set of reachable points, a dexterous workspace is admissible if all the points within it can be reached with all desired orientations. In this study, we defined the desired dexterous workspace as the locus of points reachable in every direction from the origin of the world frame such that the moving platform can rotate $\pm 30^\circ$. There are a variety of methods that can be used to determine the dexterous workspace for a given robot (e.g.\corrlab{R4.21}{,}  \cite{merlet1997democrat}). In our work, we used a hierarchical workspace scan starting with constant-orientation scans. For each constant orientation scan, a recursive depth-first search in the radial direction was used in conjunction with inverse kinematics to quickly identify the farthest reachable point associated with a particular polar coordinate. The boundary of the dexterous workspace is then determined as the set of points belonging to all the constant-orientation workspaces within the desired orientation range. This dexterous workspace is identified by \circled{1} and a dashed blue line in Fig.~\ref{fig:optimization_configuration_pic}A and B. In the case of the narrow layout, we limited the scanning range to $\pm120^\circ$ to avoid physical overlap of the base joints and the moving platform.
\par Since our goal is to offer a static balancing strategy that maintains design simplicity and compactness, we advocate for optimizing the design of the balancing mechanism for a \textit{task-based dexterous sub-workspace}. To illustrate the approach, we use the design task of following a spiral trajectory over a circular area with a diameter of $100 mm$. Once the design task has been defined, we search for all the points in the dexterous workspace that are feasible center-points for completing the design task (e.g.\corrlab{R4.21}{,} following the spiral). Points offset from the boundary of the dexterous workspace such that the robot can satisfy design task are feasible points. The boundary of such dexterous sub-workspace is identified by \circled{2} and marked by a dashed red line in Fig.~\ref{fig:optimization_configuration_pic}A and \ref{fig:optimization_configuration_pic}B.
\par
\begin{figure}[htbp]
\includegraphics[trim=0cm 6.7cm 2cm 0cm,clip, width=0.99\columnwidth]{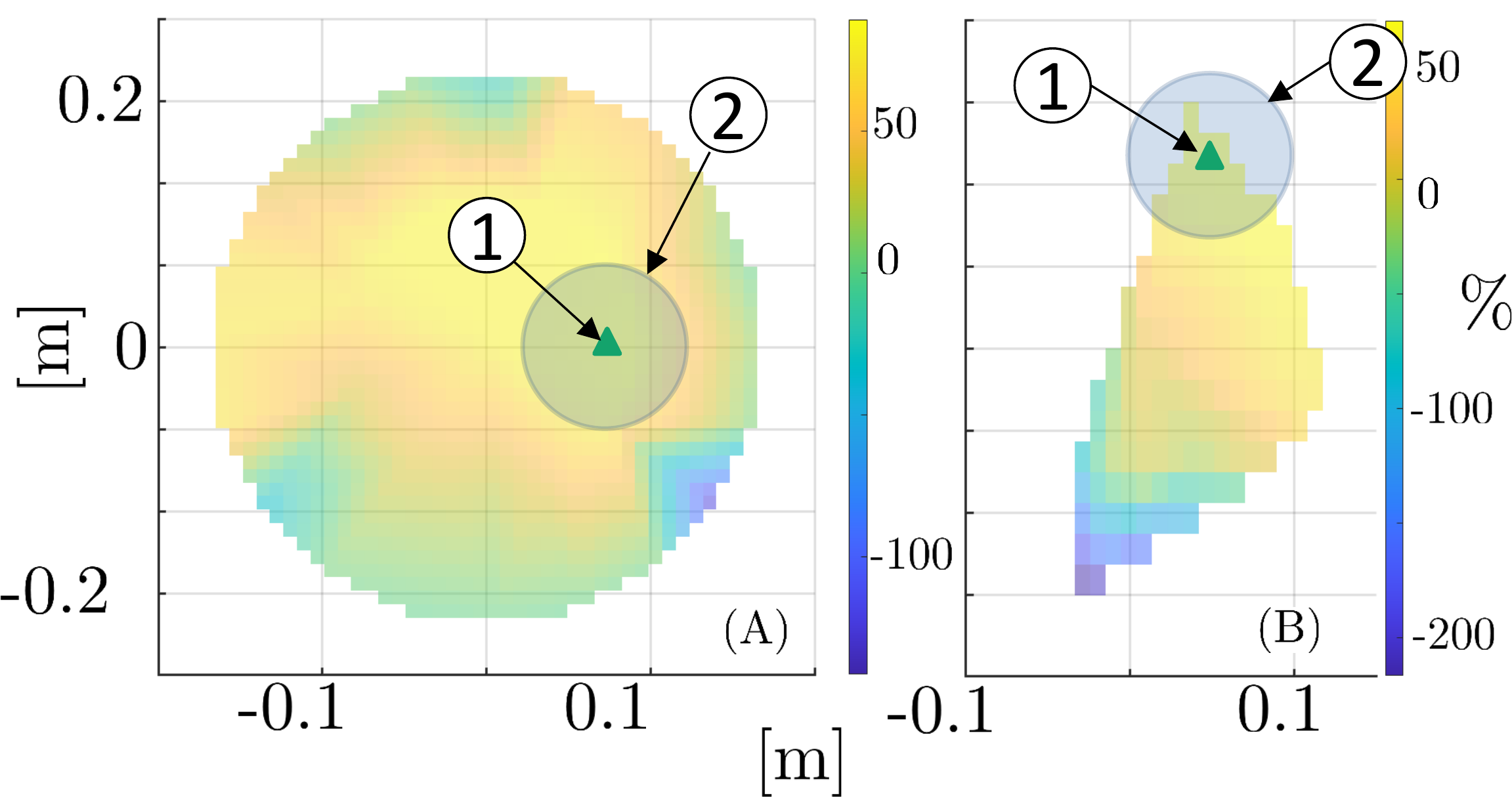}
\caption{Average percentage of torque reduction over the dexterous sub-workspace: A) WL robot configuration for $\gamma\!=\!-10^\circ$ B) NL robot configuration for $\gamma\!=\!0^\circ$.}
\label{fig:average_percent_workspace}
\end{figure}

  Figure~\ref{fig:average_percent_workspace}A-B explains why it is not feasible to use torsional springs and a design strategy that optimizes the design of the balancing springs throughout the entire dexterous sub workspace. The figure shows \corrlab{R1.8}{the sprial path followed by the robot and} the mapping of the average torque reduction percentage, using torsional springs at the active joints, in both robot base configurations and constant moving base orientation. \cutlab{R1.7}{The plots show that, while in some areas of the dexterous workspace we can obtain partial static balancing, the same set of spring parameters is not suitable for other areas where, actually, the average percentage of torque reduction has a negative value, which means that the required torque at the joints has been increased.}\corrlab{R1.7}{The plots show that, while in some areas of the dexterous workspace we can obtain partial static balancing, the same set of spring parameters is not suitable for other areas. Indeed, in some areas the average percentage of torque reduction has a negative value, i.e. the required joint torques has been increased.}
\par In order to identify the optimal location for the task-based dexterous sub-workspace, we scanned the dexterous sub-workspace for all the moving base orientations, as shown in Fig.\ref{fig:optimization_configuration_pic}A-\circled{3} and \ref{fig:optimization_configuration_pic}B-\circled{3}, and calculated the required torque at the active joints. The cyan and green asterisks in  Fig.\ref{fig:optimization_configuration_pic}A-\circled{4}-\circled{5} and \ref{fig:optimization_configuration_pic}B-\circled{4}-\circled{5} are the points with the highest and lowest required joint torque norm respectively.
\par To choose the optimal placement of the task relative to the robot we consider the actuator torque reduction due to balancing. We identify the optimal location as the point with the highest actuator torque reduction when comparing Mode 0 to Mode 1.
\par The robot configurations in Fig.\ref{fig:optimization_configuration_pic}A and \ref{fig:optimization_configuration_pic}B show the robot at the pose with the most effective counterbalancing contribution. In Fig.~\ref{fig:average_percent_workspace}A and \ref{fig:average_percent_workspace}B the location of these points are marked with a green triangle \circled{1} and the light-blue shaded circular area \circled{2} defines the associated task-based dexterous sub-workspace.
\subsection{Torsional springs optimization}
\par Once the location of the task-based sub-workspace is identified, in order to statically balance the manipulator over such area, we scan such area following a spiral path and calculate the required torque using eq.~\eqref{eq:tau_final_general}. We calculate the required actuator torques for $N$ sample points along the spiral path and we define an augmented torque vector $\breve{\bs{\tau}}=[\bs{\tau}_1\T, \ldots,\bs{\tau}_N\T]\T$. This vector is used in the calculation of the cost function $M=\frac{1}{2N}\breve{\bs{\tau}}\T\breve{\bs{\tau}}$ specifying the average norm-squared of the actuator torque. We use a nonlinear least squares algorithm to find the design parameters $\tilde{\mb{x}}$ that reduces its cost function.  Since we are using torsional springs as elastic elements to counterbalance the required torque, the design parameters vector $\tilde{\mb{x}}$ includes the initial pre-load of the springs and the stiffness coefficients for each torsional spring. The process is repeated for all the balancing modes introduced in Sec.~\ref{sec:3RRR} and the results are reported in this section.
\par
\begin{figure}[htbp]
    \centering
    \includegraphics[width=0.95\textwidth]{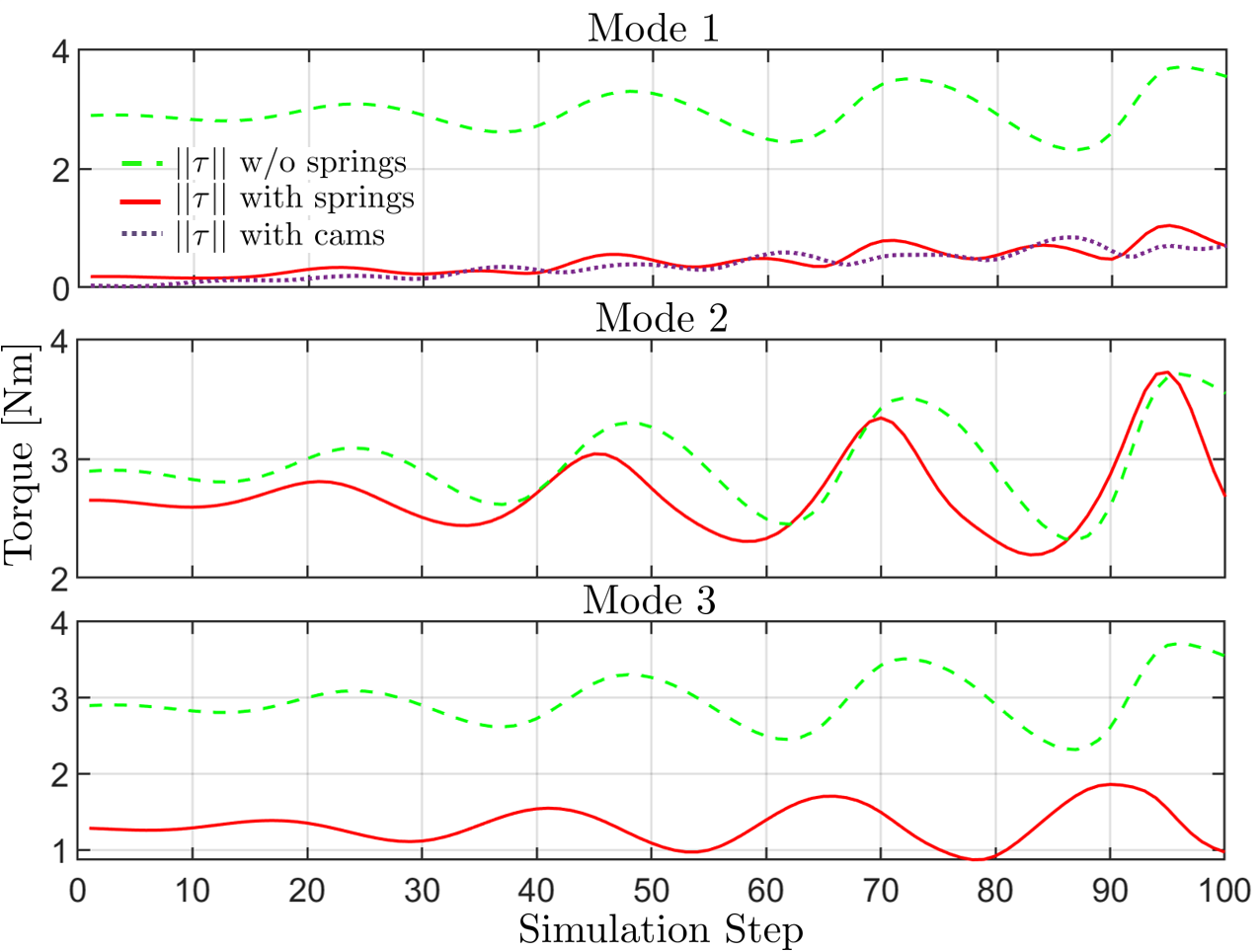}
    \caption{Torque norm profiles for WL layout.}
    \label{fig:wide_torque_profile}
\end{figure}

\begin{figure}[htbp]
    \centering
    \includegraphics[width=0.95\textwidth]{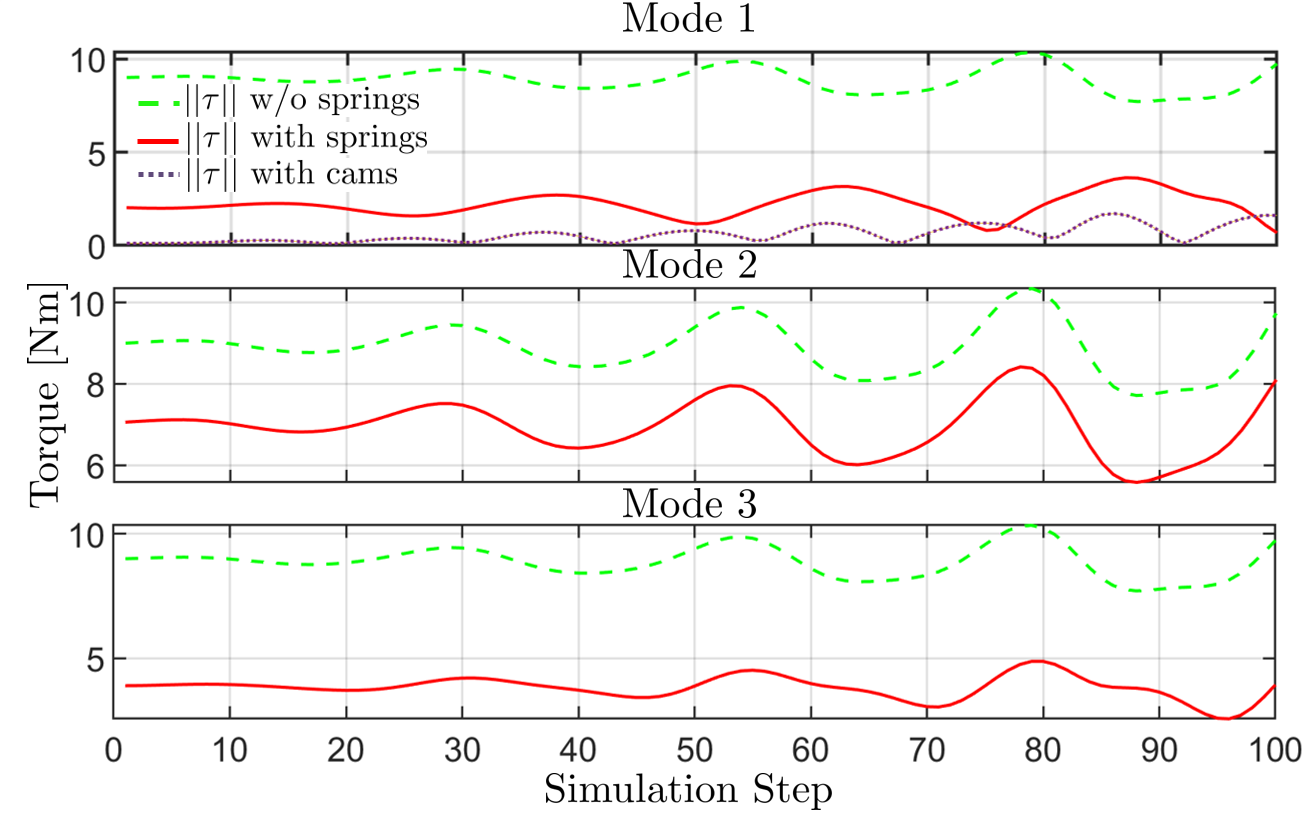}
    \caption{Torque norm profiles for NL layout.}
    \label{fig:narrow_torque_profile}
\end{figure}

 Figure~\ref{fig:wide_torque_profile} and \ref{fig:narrow_torque_profile} show the required joint torques for the WL and NL layouts over the 4 modes defined in section~\ref{sec:3RRR}. Each plot has a dashed green line which represents the norm of the actuation torques required at Mode 0, the solid red lines represent the norm of the joint torques for the three modes, finally the purple dotted lines represent norm of the torque when using the wire-wrapped cams discussed in sec.~\ref{sec:wrapped_cam}.
\par To compare the balancing efficacy of the torsional spring design, the following performance measure $ e_{\tau_i}$, was defined\corrlab{R2.11, R1.10}{}:
 \cut{\begin{equation}\label{eq:e_tau_i}
   e_{\tau_i} = \sqrt{\frac{1}{N}\sum_{1}^{N}\left(\frac{||\bs{\tau_i}_{mode_c}||}{||\bs{\tau_i}_{mode_0}||}\right)^2} \;\; c = 1,2,3
 \end{equation}}
 \corr{\begin{equation}\label{eq:e_tau_i}
   e_{\tau_i} = \sqrt{\frac{1}{N}\sum_{j=1}^{N}\left(\frac{\tau_{i,j_{mode_c}}}{\tau_{i,j_{mode_0}}}\right)^2} \;\; c = 1,2,3
 \end{equation}}
where $i$ is the index of the leg, $c$ is an index designating the Mode, and $N=1500$ is the number of points scanned in the task-based dexterous sub-workspace. \corrlab{R1.10}{Equation~\eqref{eq:e_tau_i} was designed to present the root-mean-squares of the torque ratios over the workspace. It presents the normalized reduction in torque at each location within the workspace and retains correspondence between poses for the modes being compared.} Perfect static balancing corresponds with $e_{\tau_i}=0$; therefore, a larger $e_{\tau_i}$ indicates worse balancing performance.  Table~\ref{tab:torque_percentage} reports this performance measure for all three modes. The lower values designate better static balancing outcomes for each leg.
%
\begin{table}[htbp]
\fontsize{8}{9}\selectfont
\centering
\caption{$e_{\tau_i}$ using torsional springs}\label{tab:torque_percentage}
\resizebox{0.9\columnwidth}{!}{%
\begin{tabular}{|c|c|c|c|c|c|c|c|c|}
 \hline
Mode  & \multicolumn{2}{c|}{1}                          & \multicolumn{2}{c|}{2}  & \multicolumn{2}{c|}{3}  \\ \thickhline
           & WL & NL & WL          & NL         & WL           & NL         \\ \hline
$e_{\tau_1}$   & 0.18                  & 0.12                 & 0.62       & 0.68       & 0.94        & 0.22      \\ \hline
$e_{\tau_2}$   & 0.12                 & \corr{0.9}                   & 0.70        & \corr{1.6}       & 0.08        & \corr{4.2} \\ \hline
$e_{\tau_3}$   & 0.17                 & 0.37                 & 0.85       & 0.9        & 0.29        & 0.56       \\ \hline
\end{tabular}}
\end{table}
\par
\begin{figure}[htbp]
    \centering
    \includegraphics[width=0.85\textwidth]{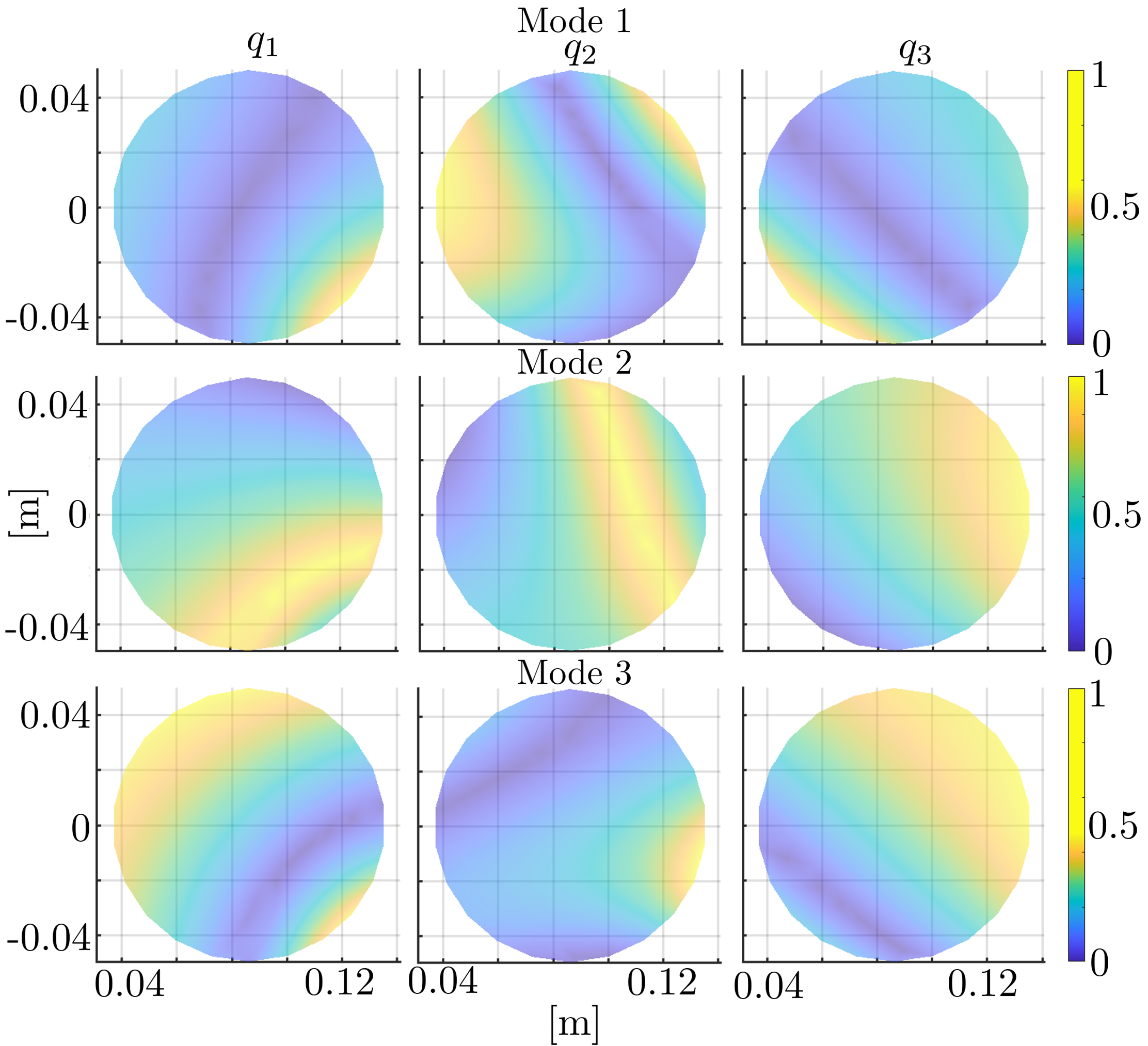}
    \caption{Torque norm ratio (with springs/without springs) for WL robot configuration}
    \label{fig:wide_q_contourf}
\end{figure}

\begin{figure}[htbp]
    \centering
    \includegraphics[width=0.85\textwidth]{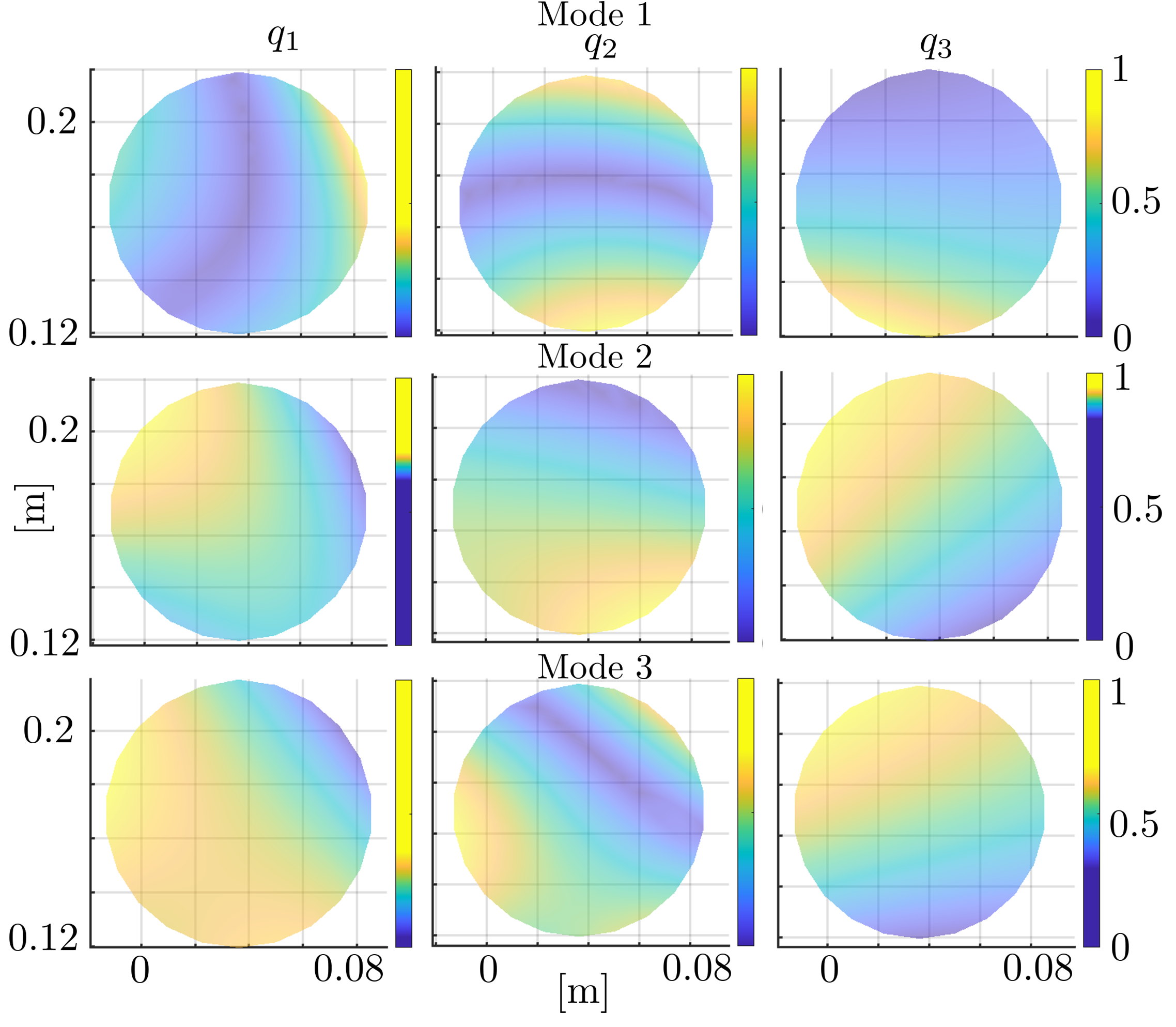}
    \caption{Torque norm ratio (with springs/without springs) for NL robot configuration}
    \label{fig:narrow_q_contourf}
\end{figure}

 Figures \ref{fig:wide_q_contourf} and \ref{fig:narrow_q_contourf} show the required \corr{absolute value torque ratio $\frac{|\tau_{{i,j}_{mode_c}}|}{|\tau_{{i,j}_0}|}$, $c=1,~2,~3$}\cut{torque norm ratio $\frac{||\bs{\tau}_||}{||\bs{\tau}_0||}$}, where $i$ designates the mode index ($i=1,2,3$) and $\bs{\tau}_0$ designates the torque without static balancing. The plots are generated over the task-based dexterous sub-workspace.

\subsection{Wire-Wrapped Cams optimization}\label{sec:wrapped_cam}
\par
\begin{figure}[htbp]
    \centering
    \includegraphics[width=0.5\columnwidth]{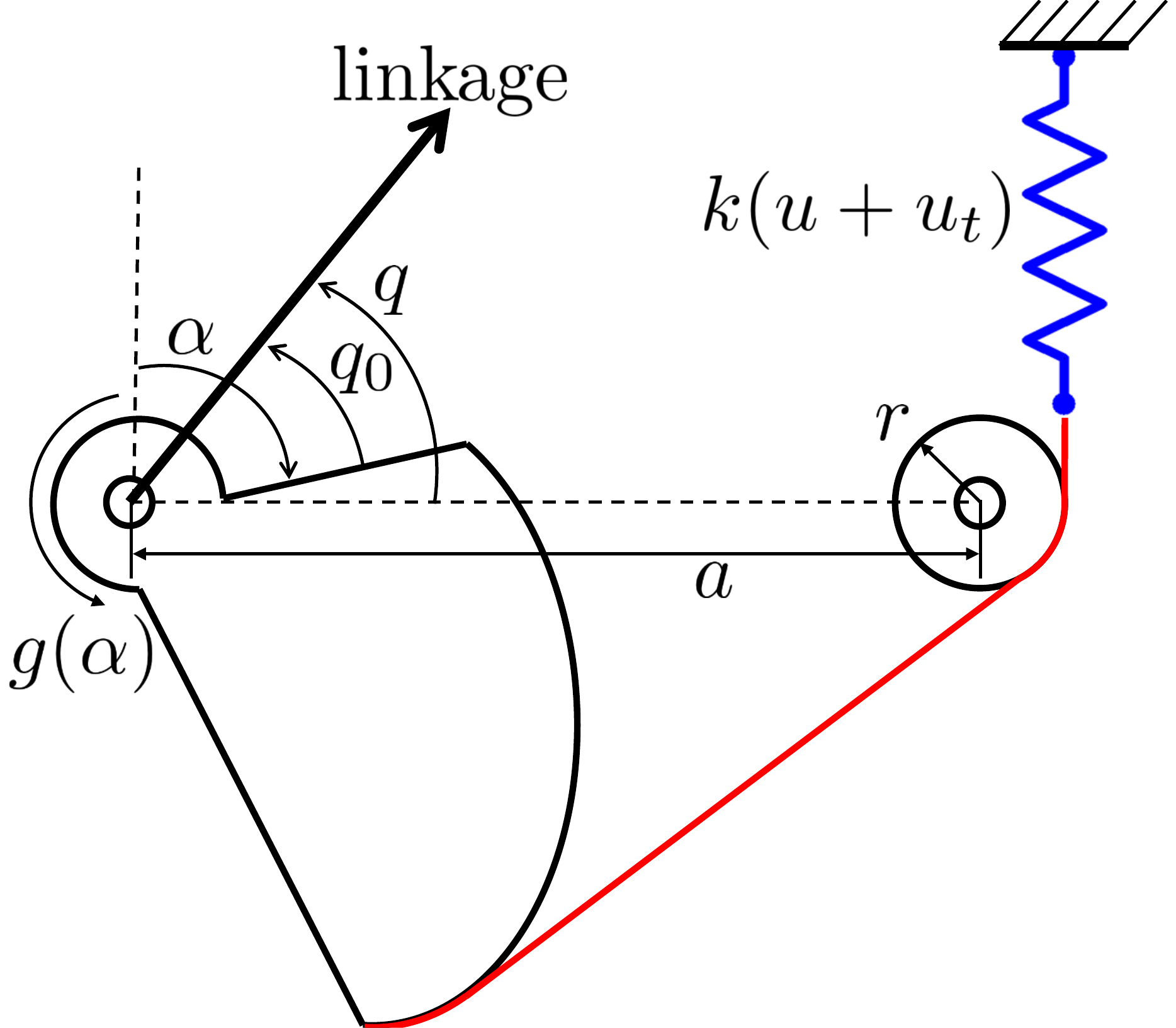}
    \caption{Cam design concept and terminology}
    \label{fig:cam_terms}
\end{figure}

 In this section, we will compare our method to a more complicated alternative that utilizes wire-wrapped cams. The wire-wrapped cam design concept we are using in this section was developed by Kilic et. al. \cite{Kilic2012}. In this method, a wire attaches on one end to a translational spring, wraps around an idler pulley, and then wraps around a cam as shown in Fig. \ref{fig:cam_terms}. When the cam rotates, it extends the spring causing a moment about the cam's axis of rotation. Using \cite{Kilic2012}, the cam can be specially designed to generate a desired moment $g(\alpha)$ as a function of cam angle $\alpha$. In the design shown in Fig. \ref{fig:cam_terms}, the idler and cam centers of rotation are assumed to be horizontally aligned and separated by a distance $a$ along the $x$ axis. The designer is free to choose $a$ as a design parameter. The translational spring constant $k$, the spring pre-extension $u_t$, and the idler radius $r$ are also design parameters.
\par
\begin{figure}[htbp]
    \centering
    \includegraphics[width=0.5\textwidth]{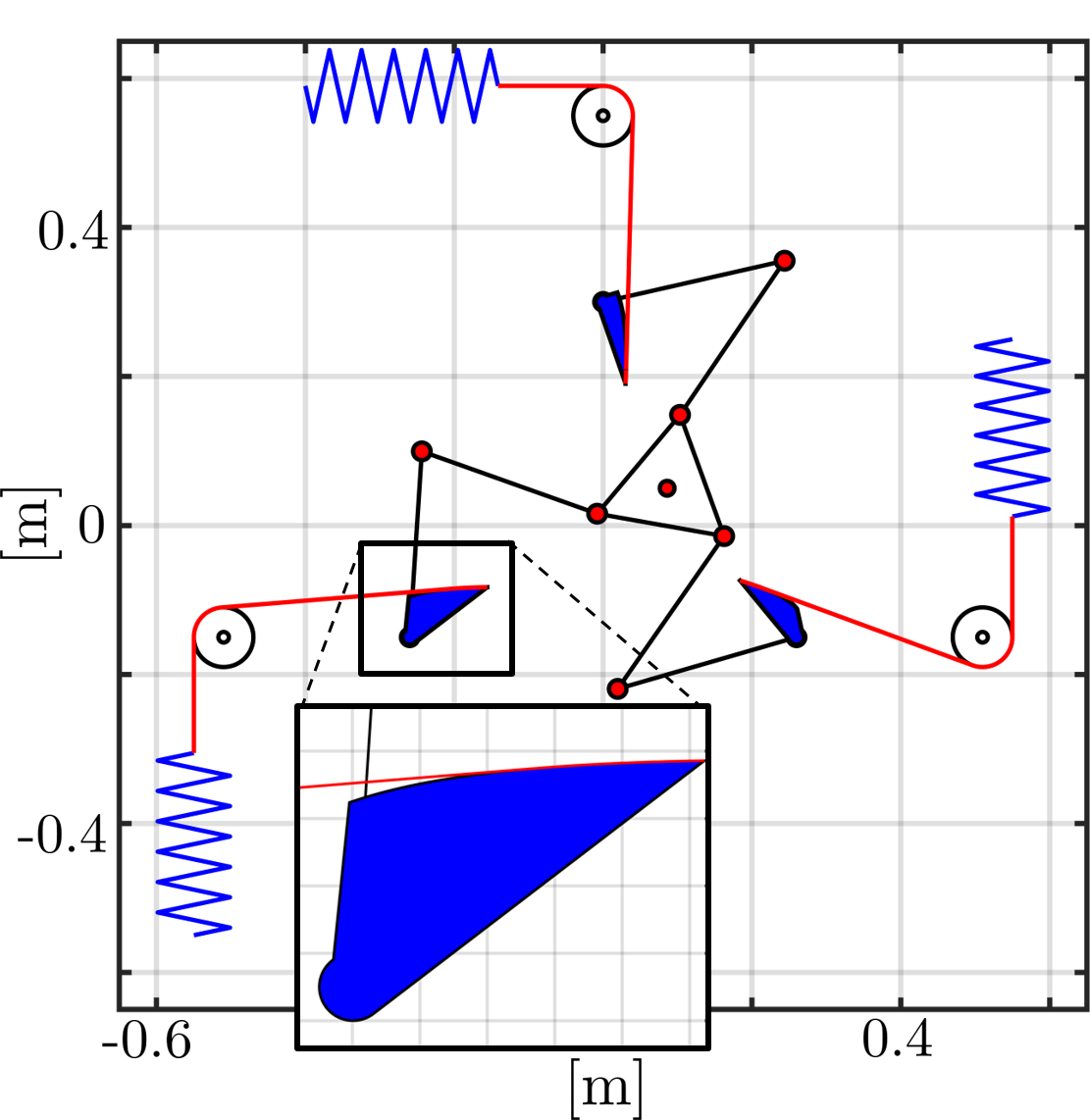}
    \caption{3RRR wire-wrapped cam design concept for the WL case}
    \label{fig:cam_3rrr}
\end{figure}

 By rigidly attaching a cam to each active joint in the 3RRR robot as shown in Fig. \ref{fig:cam_3rrr}, we can offset the gravitational torque on the joints in a section of the workspace.
 \par The equations in \cite{Kilic2012} require the cam angle $\alpha$ to be defined as shown in Fig. \ref{fig:cam_terms}. Because $\alpha$ is defined differently from the linkage angle $q$, we must calculate the cam angle for a given linkage angle using:
 \begin{equation}
   \alpha = \frac{\pi}{2} - (q - q_0)
 \end{equation}
 Where the constant angle between the cam and the linkage $q_0$ is a design parameter. Because the torque on each joint from the wire-wrapped cam is a function only of the angle of the joint it is attached to, we can only achieve approximate balancing. Therefore, we must design a function $\tau_{est}(\alpha)$ that is the best possible estimate of $\tau$. To do this, we chose an $n^\text{th}$ order polynomial basis for $\tau_{est}(\alpha)$:
 \begin{equation}\label{eq:tau_modal}
   \tau_{est,i}(\alpha_i) = b_{i,0}+b_{i,1}\alpha_i+b_{i,2}\alpha_i^2+\dots+b_{i,n}\alpha_i^n\quad i=1,2,3
 \end{equation}
 \noindent In this equation, $i$ represents the joint (and cam) number. \corrlab{R1.11}{Although one could have chosen Bernstein or Chebyshev polynomials for their favourable numerical conditioning for higher powers, we chose a simple monomial basis since the maximal power was $n=4$. This monomial basis remains a reasonable choice since it was shown in \cite{Angeles2012_cam_optimization} that it becomes numerically poorly conditioned for $n>7$}.
 \par
\begin{figure}[htbp]
    \centering
    \includegraphics[width=0.99\textwidth]{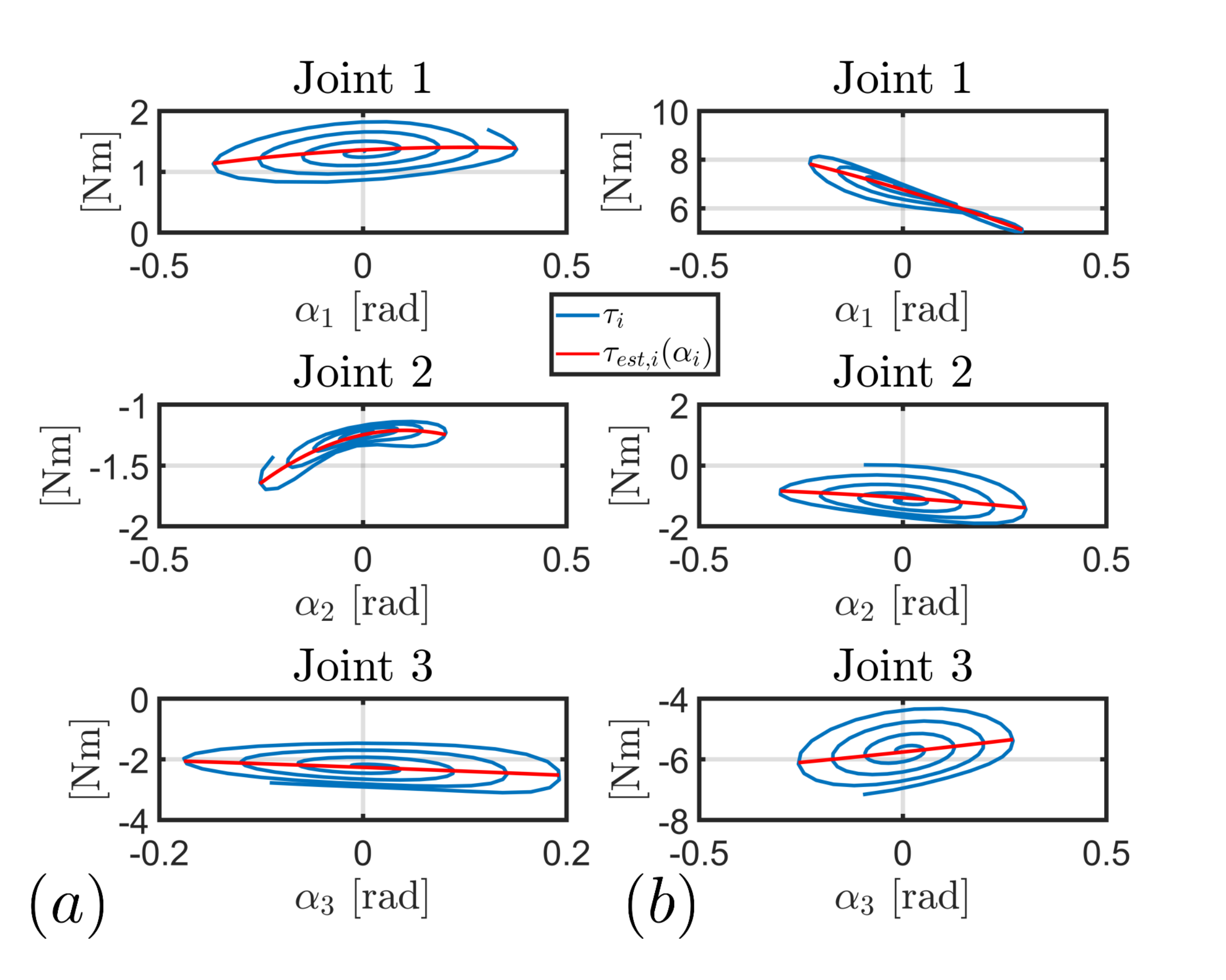}
    \caption{Actual torque $\tau_i$ vs. least squares torque estimate $\tau_{est,i}$: (a) WL case (b) NL case}
    \label{fig:torque}
\end{figure}

  We now must choose the modal coefficients $\mb{b}_i = [b_{i,0}, b_{i,1}, \dots,b_{i,n}]\T\in\realfield^{n+1}$ that minimize the difference between $\tau_{est,i}(\alpha_i)$ and the actual torque on the joint $\tau_i$. If we \rewrite{calculate}\cut{take measurements of} $\tau_i$ at $N$ different poses of the robot, we can write the following equation:
 \begin{equation}
   \underbrace{\begin{bmatrix}
     \tau_{i,1} \\
     \vdots \\
     \tau_{i,N}
   \end{bmatrix}}_{\bs{\tau}_i} = \underbrace{\begin{bmatrix}
                                                1 & \alpha_{i,1} & \alpha_{i,1}^2 & \dots & \alpha_{i,1}^n\\
                                                {}& {}           & \vdots         & {}    & {}\\
                                                1 & \alpha_{i,N} & \alpha_{i,N}^2 & \dots & \alpha_{i,N}^n
                                              \end{bmatrix}}_{\mb{M}_i}\underbrace{\begin{bmatrix}
                                                                         b_{i,0} \\
                                                                         b_{i,1} \\
                                                                         \vdots \\
                                                                         b_{i,n}
                                                                       \end{bmatrix}}_{\mb{b}_i}, \quad i = 1,2,3
 \end{equation}
\noindent We can now solve for the modal coefficients using the pseudo-inverse of $\mb{M}_i$:
 \begin{equation}\label{eq:modal_coeffs}
   \mb{b}_i = \mb{M}_i^+\bs{\tau}_i,\quad i = 1,2,3
 \end{equation}
Because we are trying to balance $\tau$, the desired cam torque as a function of cam angle can be found using:
 \begin{equation}
   g_i(\alpha_i) = -\tau_{est,i}(\alpha_i),\quad i = 1,2,3
 \end{equation}
 For the both the WL and NL cases, a $\tau_{est,i}(\alpha_i)$ function was calculated from \corr{\eqref{eq:tau_modal} and} \eqref{eq:modal_coeffs} using the data from the search inside the task-based sub-workspace corresponding with the optimal location for Mode 1. The result of this analysis is shown in Fig. \ref{fig:torque}.
\par Now that the desired torque function $g_i(\alpha_i)$ has been calculated for each cam, we can design the cams using the parameters shown in Table \ref{tab:cam_const}. The resulting cams for the WL case are shown in Fig. \ref{fig:cam_3rrr}.
\begin{table}[htbp]
 \fontsize{9}{10}\selectfont
\centering
\caption{Cam Design Constants}\label{tab:cam_const}
\resizebox{0.9\columnwidth}{!}{%
\begin{tabular}{|c|c|c|c|c|c|}
\hline
\multicolumn{2}{|c|}{}        & $q_0$       & $a$    & $u_t$    & $r$\\ \hline
\multirow{3}{*}{WL}   & $q_1$ & 2.1567 rad  & 0.25 m   & 0.05 m  & 0.04 m\\
                      & $q_2$ & -1.5669 rad & 0.25 m   & 0.065 m & 0.04 m\\
                      & $q_3$ & -0.1566 rad & 0.25 m   & 0.05 m  & 0.04 m\\ \hline
\multirow{3}{*}{NL}   & $q_1$ & -0.2287 rad & 0.0414  & 0.06 m  & 0.04 m\\
                      & $q_2$ & 0.2582 rad  & 0.0414   & 0.05 m  & 0.04 m\\
                      & $q_3$ & 0.0019 rad  & 0.0414  & 0.06 m  & 0.04 m\\ \hline
					
\end{tabular}}
\end{table}
\par \corrlab{R2.11}{The required motor torque for leg $i$ at position $j$ with cam balancing is given by the sum $g_i(\alpha_{i,j})+\tau_{i,j}$ and without balancing is given by $\tau_{i,j}$. We therefore express \eqref{eq:e_tau_i} for this case the following in order to compare the cam balancing efficacy of the wire-wrapped cam designs:
 }
\par \cut{Similarly to the torsional spring case, eq.\eqref{eq:e_tau_i} is used for the wire-wrapped cam solution to compare the cam balancing efficacy of the wire-wrapped cam designs, the following performance measure was created:}
\begin{equation}\label{eq:e_tau_i_cam}
  e_{\tau_i} = \sqrt{\frac{1}{N}\sum_{j=1}^{N}\left(\frac{g_i(\alpha_{i,j})+\tau_{i,j}}{\tau_{i,j}}\right)^2}\quad i = 1,2,3
\end{equation}
\par
\begin{figure}[htbp]
    \centering
    \includegraphics[width=0.99\textwidth]{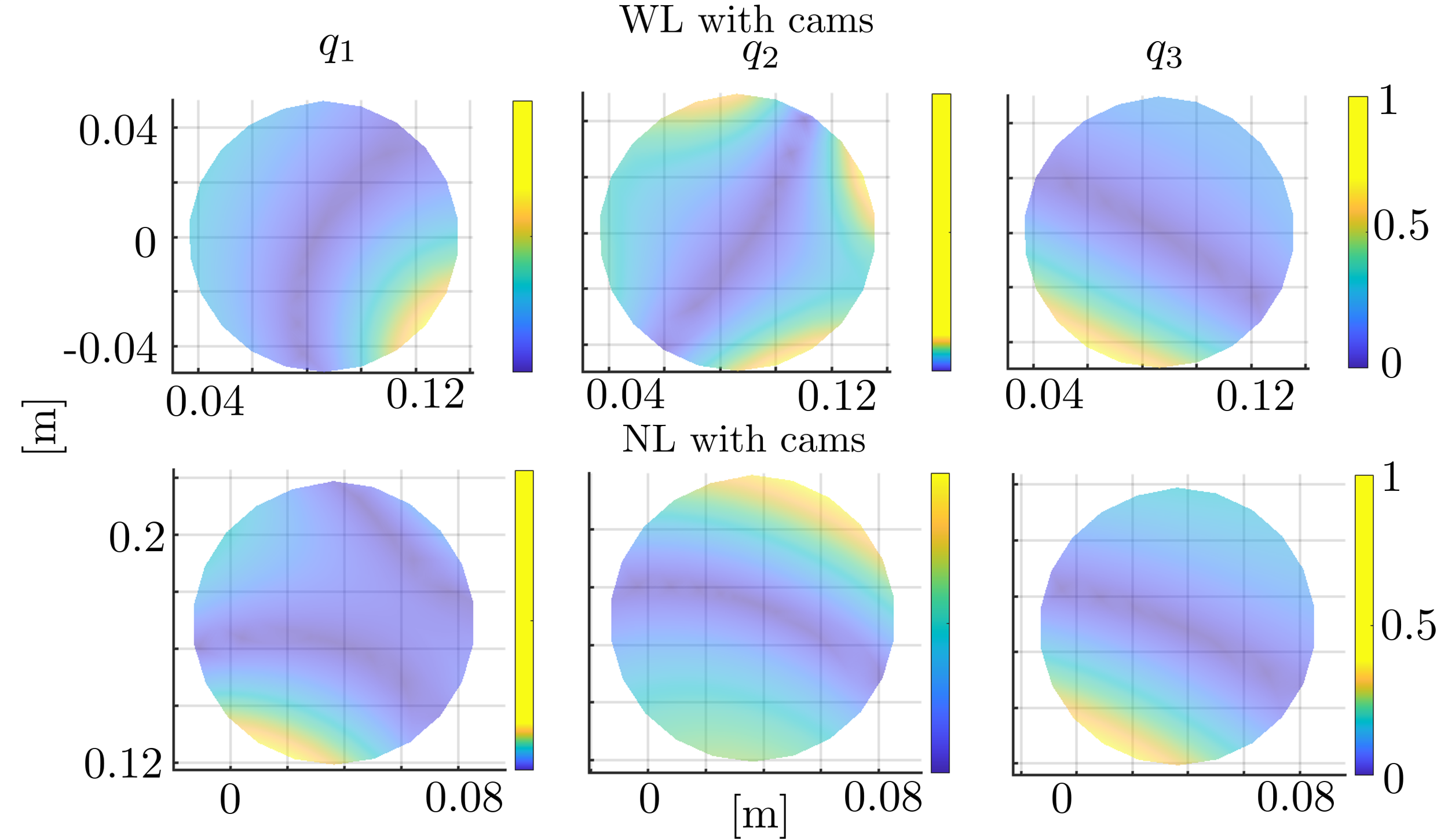}
    \caption{Torque norm ratio (with springs/without springs) for WL and NL layouts using wire-wrapped cams.}
    \label{fig:cam_contourf}
\end{figure}

 Table \ref{tab:measures} shows the values of $e_{\tau_i}$ in the case of wire-wrapped cams. Figure~\ref{fig:cam_contourf} shows the torque norm ratio $\frac{||\bs{\tau}_{with~cams}||}{||\bs{\tau}_0||}$ over the task-based dexterous sub-workspace.
%
%

\begin{table}[htbp]
 \fontsize{8}{9}\selectfont
\centering
\caption{$e_{\tau_i}$ using cams}\label{tab:measures}
\resizebox{0.5\columnwidth}{!}{%
\begin{tabular}{|c|c|c|c|}
\hline
{}                                 & $e_{\tau_1}$      & $e_{\tau_2}$ & $e_{\tau_3}$\\  \thickhline
  WL& 0.17  & 0.045 & 0.16  \\
                      NL  & 0.042  & 0.39& 0.11  \\ \hline
					
\end{tabular}}
\end{table}

\section{Discussion}
The simulation study informs the answers to the questions we posed at the outset when embarking on this study:
\arrowbullet \textbf{Effect of layout:} The simulation results show that the WL layout is easier to balance than the NL layout as indicated by the lower performance measures $e_{\tau_i}$. Also, the worst performance corresponds with the kinematic \cutlab{R1.12}{cain}\corrlab{R1.12}{chain} that reaches the biggest extension. 
\arrowbullet \textbf{Effect of balancing torsional spring location:} The comparison of modes shows that Mode 1 offers the best performance based on Table~\ref{tab:torque_percentage}. Mode 2 offers the worst balancing performance and Mode 3 does not improve over Mode 1. Therefore, if one limits the design implementation to using torsional springs, these springs should be located at the base of the robot where the active joints are positioned.
\arrowbullet \textbf{Wire-wrapped cams v.s. torsional springs:} When comparing the wire-wrapped cam design to the torsional spring design with Mode 1, we observe by comparing the first two columns of Table~\ref{tab:torque_percentage} to the corresponding two rows of Table~\ref{tab:measures} that the cam-based balancing marginally improves balancing performance for the WL layout, but offers significant improvement for the NL layout. 
\par Based on these results, if one chooses a WL layout based on an application compatible with this layout, then using a simple balancing solution based on torsional springs can offer significant actuator torque reduction over a task-based workspace. Alternatively, if an application demands the use of the NL layout and the design size limitations can admit the added size and complexity associated with the use of idlers, cams, and linear springs then the cam-based design will offer better balancing performance. If however the size limitations are severe then the use of a simple design with torsional springs can offer effective static balancing over a desired task-based workspace.
\par For the simple design alternative using torsional springs, changing the location of the dexterous workspace requires redesign of the spring preload and spring constants and, in some aspects of the workspace, may not produce effective balancing as shown in Fig.~\ref{fig:average_percent_workspace}.The cam-based design alternative effectively locks the location of the dexterous workspace since it requires a change in design to incorporate a clutch mechanism allowing the cams to disengage while shifting the center of the desired workspace. Such mechanism will ensure that the cams still operate within the range they were designed for. However, the static balancing performance in that case also would not be guaranteed to offer effective balancing.
\section{Acknowledgement}
This research was partly funded by NSF award \#1734461 through support for G. Johnston and N. Simaan. Any opinions, findings, and conclusions or recommendations expressed in this material are those of the authors and do not necessarily reflect the views of the National Science Foundation.
\section{Conclusion}
This paper discussed design alternatives for static balancing for parallel mechanisms using elastic elements consistent with applications demanding design compactness, low mass, and simplicity (e.g.\corrlab{R4.21}{,} as is the case for medical robotics). While it is possible to obtain perfect balancing along a path, it is hard to obtain exact balancing throughout a workspace if one wishes to avoid the use of balancing masses and to maintain design simplicity. We have considered two design layouts for the 3RRR parallel robot and considered the efficacy of static balancing based on the location of torsional balancing springs along the kinematic chains of each design layout. We also compared the performance of these simple designs to a design alternative using wire-wrapped cams. Our design strategy explains how to use the concept of a task-based workspace for the optimal design of balancing spring parameters and for the choice of the workspace location. By using modal fitting for the required torque throughout the task-based design workspace, we are able to adapt recent results for the design of wire-wrapped cams for the balancing of the two possible design layouts. The methodology of this paper provides a useful guidance based on the constraints of the application scenario in terms of choosing the design layout, the placement of torsional springs, or the possible use of wire-wrapped cams for balancing.


\bibliographystyle{asmeconf}
\bibliography{bib/static_balancing,bib/giuseppe_bibliography}
\balance
\end{document}